\documentclass{article}
\usepackage{ijcai24}

\usepackage{times}
\usepackage{soul}
\usepackage{url}
\usepackage[hidelinks]{hyperref}
\usepackage[utf8]{inputenc}
\usepackage[small]{caption}
\usepackage{graphicx}
\usepackage{subcaption}
\usepackage{amsmath}
\usepackage{amssymb}
\usepackage{amsthm}
\usepackage[table]{xcolor}
\usepackage{booktabs}
\usepackage{algorithm}
\usepackage{algorithmic}
\usepackage[switch]{lineno}

\title{DVPE: Divided View Position Embedding for Multi-View 3D Object Detection}

\author{
Jiasen Wang$^1$
\and
Zhenglin Li$^{1}$\thanks{Corresponding author.}\and
Ke Sun$^{1}$\and
Xianyuan Liu$^2$\And
Yang Zhou$^{1}$
\affiliations
$^1$Shanghai University \\
$^2$University of Sheffield\\
\emails
\{wangjiasen, zhenglin\_li, ke\_sun\}@shu.edu.cn,
xianyuan.liu@sheffield.ac.uk,
saber\_mio@shu.edu.cn
}

\begin{document}

\maketitle

\begin{abstract}
    Sparse query-based paradigms have achieved significant success in multi-view 3D detection for autonomous vehicles. Current research faces challenges in balancing between enlarging receptive fields and reducing interference when aggregating multi-view features. Moreover, different poses of cameras present challenges in training global attention models. To address these problems, this paper proposes a divided view method, in which features are modeled globally via the visibility cross-attention mechanism, but interact only with partial features in a divided local virtual space. This effectively reduces interference from other irrelevant features and alleviates the training difficulties of the transformer by decoupling the position embedding from camera poses. Additionally, 2D historical RoI features are incorporated into the object-centric temporal modeling to utilize high-level visual semantic information. The model is trained using a one-to-many assignment strategy to facilitate stability. Our framework, named DVPE, achieves state-of-the-art performance (57.2\% mAP and 64.5\% NDS) on the nuScenes test set. Codes will be available at \url{https://github.com/dop0/DVPE}.
\end{abstract}

\begin{figure*}[t]
	\centering
    \begin{subfigure}[b]{0.28\textwidth}
        \includegraphics[width=\textwidth]{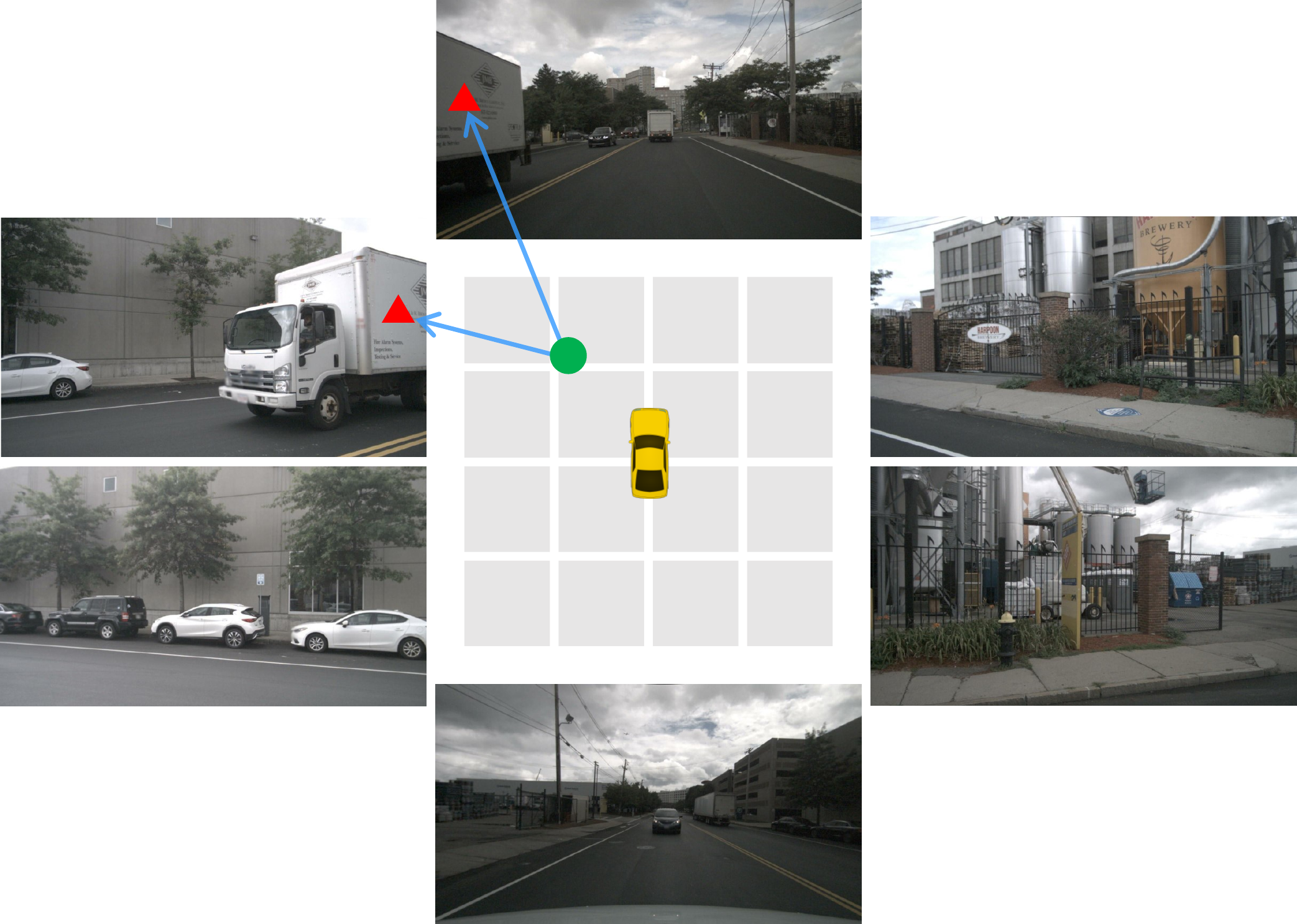}
        \caption{\textbf{Sparse Sampling Method}}
        \label{fig:sparse_sampling_method}
    \end{subfigure}
    \hfill
    \begin{subfigure}[b]{0.28\textwidth}
        \includegraphics[width=\textwidth]{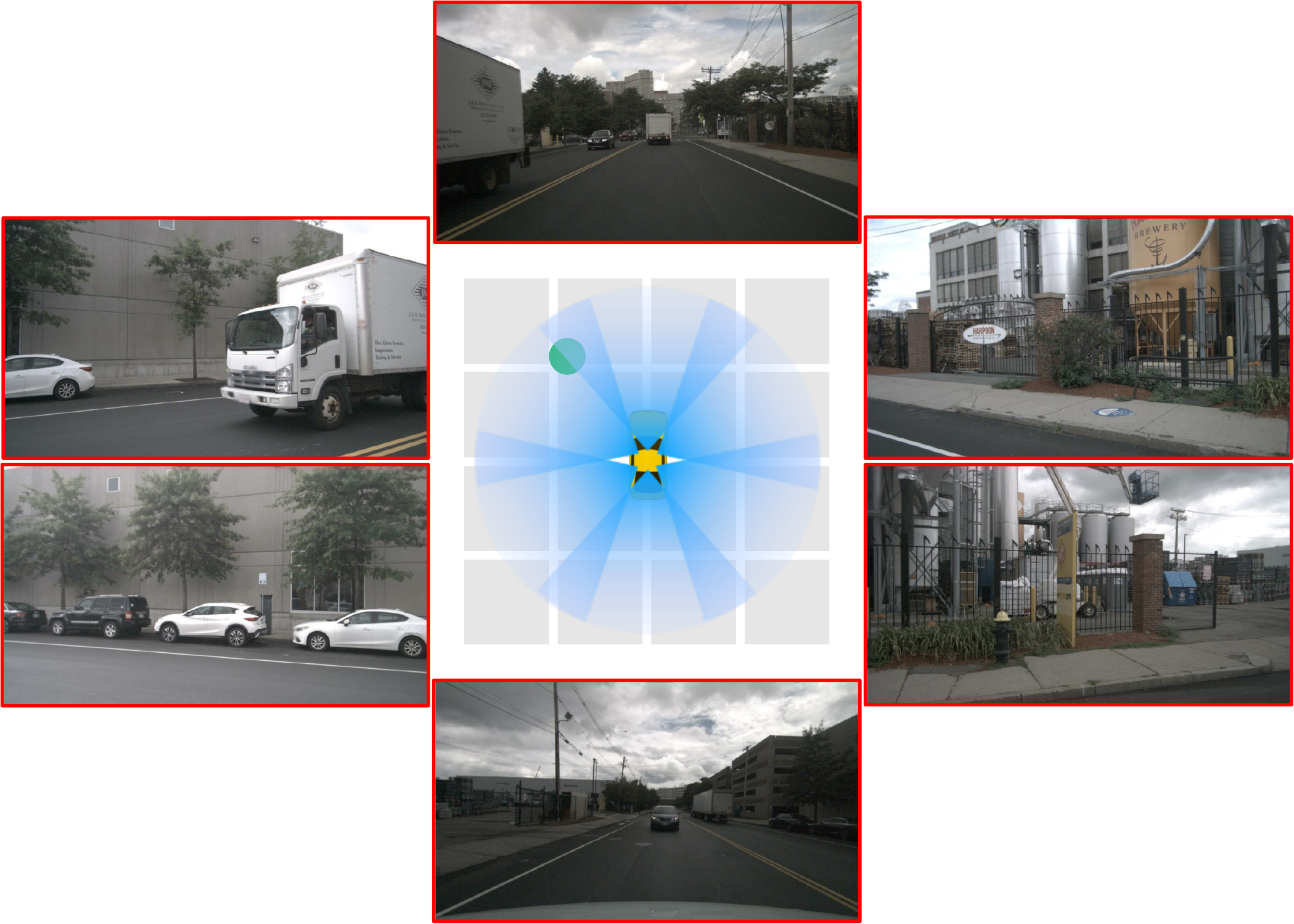}
        \caption{\textbf{Global Attention Method}}
        \label{fig:global_attention_method}
    \end{subfigure}
    \hfill
    \begin{subfigure}[b]{0.25\textwidth}
        \includegraphics[width=\textwidth]{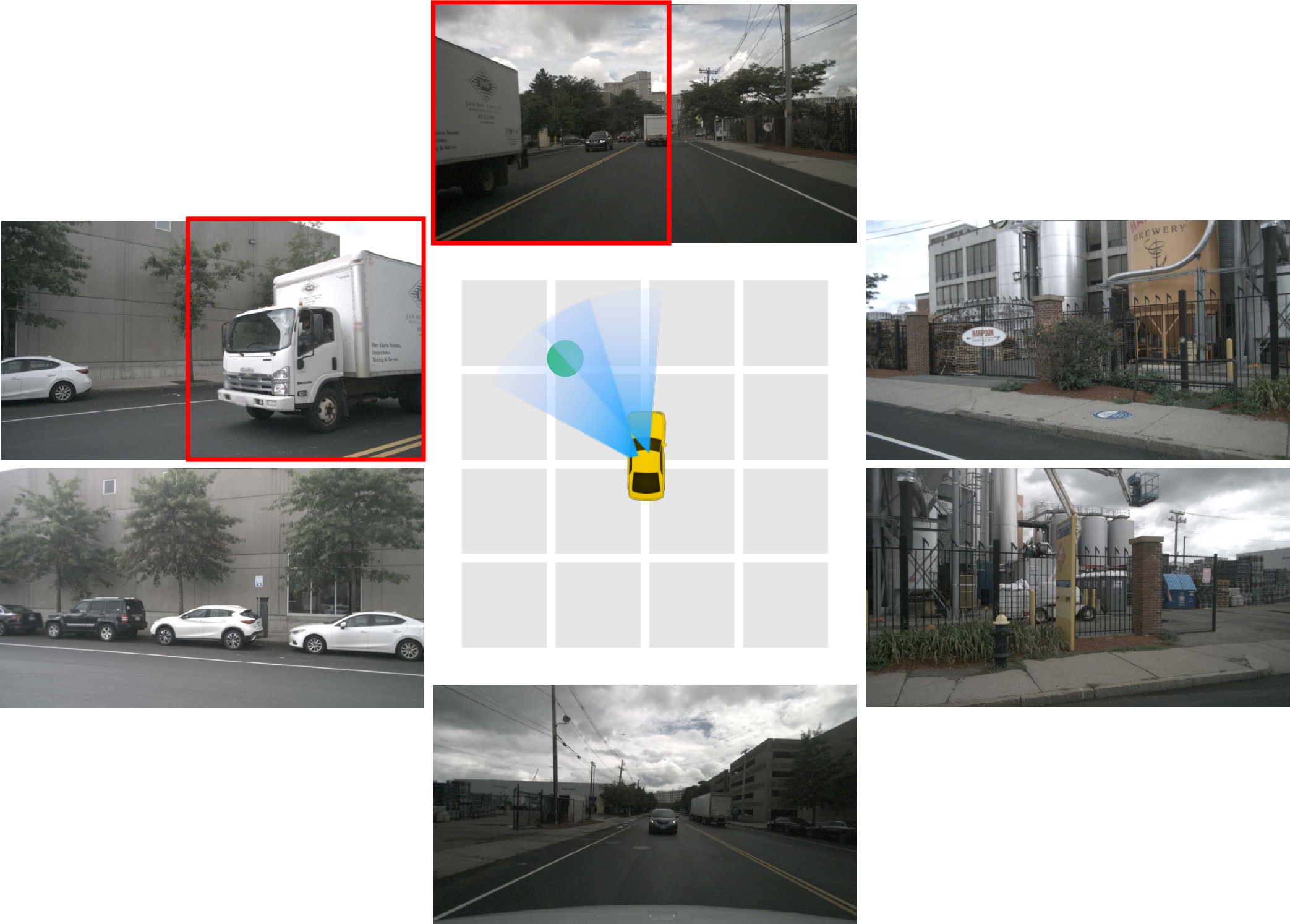}
        \caption{\textbf{Divided View Method (ours)}}
        \label{fig:divided_view_method}
    \end{subfigure}
    \caption{Illustration of the image feature aggregation process of sparse sampling, global attention, and our proposed divided view approach, all belonging to sparse query-based paradigms.
    The green dot represents the 3D reference point of a query. 
    Solid lines indicate projection onto the image plane, while triangles represent projected sampling points on images. 
    The portion enclosed by a red frame represents the image features that need to interact with queries through cross-attention. 
    The blue frustums represent the regions confirmed by camera rays, where the 3D coordinates are encoded to position embedding for cross-attention operations.}
    \label{fig:introduction}
\end{figure*}

\section{Introduction}

Recently, vision-based multi-view 3D object detection in autonomous driving has garnered considerable attention due to its low cost and remarkable performance. 
Existing research can be categorized into two main groups:
Bird's Eye View (BEV)-based and sparse query-based paradigms. 
The former~\cite{philion2020lift,li2022bevformer,li2023bevdepth}
conducts view transformation to explicitly construct BEV 
feature maps from multi-view images, providing a unified intermediate representation for various downstream tasks. 
However, achieving a trade-off between perception capabilities and computational complexity proves challenging due to the dense representation of BEV features and intricate view transformation. 
Comparatively, sparse query-based methods~\cite{wang2022detr3d,liu2022petr} typically predefine sparse query embedding or reference points in 3D space and iteratively aggregate image features, which has been proven effective in 3D object detection. 

Existing sparse query-based methods can be summarized into two branches based on the way of aggregating image features: sparse sampling and global attention. 
Sparse sampling-based approaches extract features through linear interpolation at specific projected points within image feature maps, as illustrated in Figure~\ref{fig:sparse_sampling_method}. 
DETR3D~\cite{wang2022detr3d} pioneered this group of methods, which projects 3D reference points onto vision feature maps based on camera parameters to sample features. 
While sparse sampling is efficient in terms of inference speed, the limited number of sampling points makes it challenging to achieve accurate global modeling, thereby hindering optimal performance. 
PETR~\cite{liu2022petr}, as a representative work of global attention-based approaches (as illustrated in Figure~\ref{fig:global_attention_method}), 
encodes 3D coordinates as position embedding based on camera rays, thereby enabling image features to become 3D position-aware. 
Subsequently, global cross-attention is performed between queries and multi-view image features, which is largely redundant. 
For instance, looking up image features of the rear view does not contribute to detecting targets in front, but will lead to interference from irrelevant objects and unnecessary computation. 
Moreover, the existing PETR series~\cite{liu2022petr,liu2023petrv2,wang2023exploring}, 
initializes 3D reference points and outputs predictions in the 3D world space (i.e., the ego coordinate system).
This compels models to adapt queries to account for camera poses through position embedding, thereby heightening the complexity of learning. 

To address the aforementioned challenges, we propose a novel end-to-end framework based on divided view position embedding (DVPE) encoded in local virtual spaces. 
In contrast to typical position embedding methods that encode 3D coordinates of camera rays in the global world space, the proposed approach divides the 3D world space into multiple local virtual spaces.
By transforming both 3D reference points and camera 3D coordinates into these virtual spaces, it decouples position embedding from camera poses and positions, thereby reducing the model's learning complexity.  
Subsequently, as illustrated in Figure~\ref{fig:divided_view_method}, we introduce a visibility attention module, performing cross-attention separately within each virtual space. 
This enables efficient global modeling by only interacting with  
image features 'visible' to each query, instead of the entire 360° view, thus effectively avoiding interference from irrelevant features. 

Additionally, we introduce an object-centric temporal modeling approach enhanced by 2D vision prior. 
Specifically, historical Region of Interest (RoI) features from a 2D detector are incorporated to enrich the contextual information with high-level semantic features beyond decoder embedding, thereby facilitating the detection performance. 
Moreover, we explore a one-to-many assignment strategy in the training of 3D multi-view object detection frameworks, effectively enhancing the training stability by enriching the supervision of the decoder. 
The main contributions of this paper can be summarized as follows:
\begin{itemize}
\item We propose a multi-view 3D object detection framework 
named DVPE. 
It divides the global world 3D space into several local virtual spaces and performs visibility cross-attention based on the divided view position embedding, which effectively reduces the interference from irrelevant features and mitigates the learning difficulty by decoupling position embedding from camera poses.
    
\item A 2D visual information-enhanced object-centric temporal modeling approach is proposed, which caches historical instance-level information from 2D detectors to enrich 3D features for temporal fusion. 
    
\item We introduce a one-to-many assignment training strategy to 
the framework for 3D object detection, effectively stabilizing the training of the decoder.
    
\item Our model has been extensively evaluated on the nuScenes dataset, demonstrating superior performance over other camera-based 3D object detection methods and achieving state-of-the-art (SOTA) 
performance of 57.2\% mAP and 64.5\% NDS.
\end{itemize}

\begin{figure*}[t]
	\centering
	\includegraphics[width=0.77\linewidth]{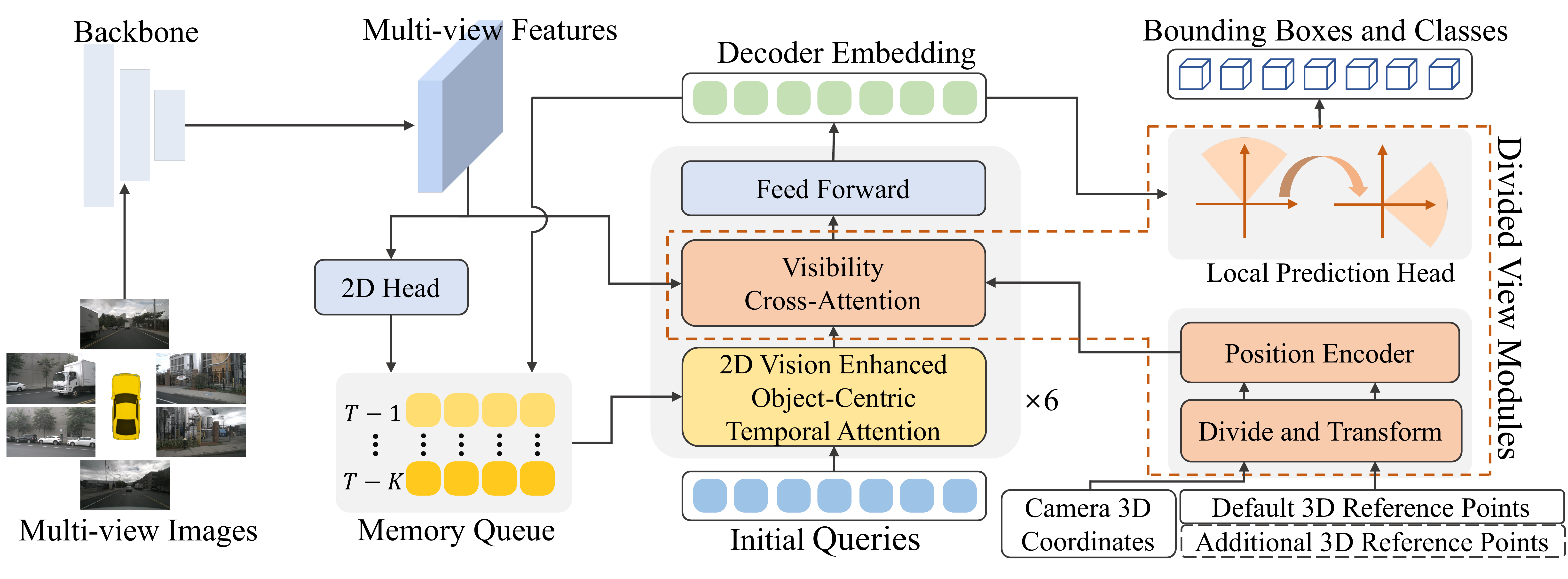}
	\caption{\textbf{Overall architecture of DVPE.}  The framework is based on 
    the transformer decoder, where the initial queries update iteratively 
    through temporal attention and visibility cross-attention. 
    In temporal attention, object queries interact with themselves as well as historical decoder embedding and 2D RoI embedding stored in the memory queue. 
    Before visibility cross-attention, object queries and image features are grouped based on their 3D coordinates and then transformed into several local virtual spaces to obtain divided view position embedding. 
    Isolated cross-attention is performed between queries and image features within different spaces. 
    Subsequently, predictions are made in local virtual spaces and then transformed back to the 3D world coordinate system as final detection results, following which the memory queue is updated. 
    Additional 3D reference points are used in conjunction with default 3D reference points for one-to-many assignment during training.} 
	\label{fig:framework_architecture}
\end{figure*}

\section{Related Work}

\subsection{Transformer-Based Object Detection}
DETR~\cite{carion2020end} represents a pioneering work in applying the transformer decoder~\cite{vaswani2017attention} for object detection, where learnable queries interact with both image features and object information. 
Each query generates one prediction and is assigned one-to-one to the ground truth, thereby avoiding the need for non-maximum suppression (NMS). 
Many subsequent works~\cite{meng2021conditional,liu2021dab} focus on accelerating the convergence speed of DETR training. 
Specifically, Deformable DETR~\cite{zhu2020deformable} proposes a deformable attention module to localize sparse features. 
DN-DETR~\cite{li2022dn} and DINO~\cite{zhang2022dino} mitigate the instability of bipartite matching by introducing a denoising training method. 
Recent studies~\cite{jia2023detrs,chen2023group,chen2023group} achieve notable performance by introducing a one-to-many assignment method to enhance training efficiency. 

\subsection{Multi-View 3D Object Detection}

In contrast to monocular 3D object detection methods~\cite{wang2021fcos3d,wang2022probabilistic}, multi-view detection frameworks perceive multiple images from different cameras and predict bounding boxes in 3D world space by fully utilizing the correlation of multi-view images and geometric information of surrounding cameras. 
Based on whether vision features are transformed into BEV, these methods can be broadly categorized into two branches: BEV-based paradigm and sparse query-based paradigm.

\paragraph{BEV-based Paradigm.}
LSS~\cite{philion2020lift} lifts multi-view features into 3D space through depth estimation and splats them into BEV representation using voxel pooling. 
BEVDepth~\cite{huang2021bevdet} improves LSS by conducting depth supervision via LiDAR point cloud projection. 
BEVStereo~\cite{li2023bevstereo} and VideoBEV~\cite{han2023exploring} perform BEV temporal fusion to enhance the performance.
Different from LSS, many frameworks project image features back to BEV, typically using a transformer decoder.
BEVFormer~\cite{li2022bevformer} predefines a grid of reference points in BEV and projects them onto images in PV to obtain corresponding features with a transformer.
PolarFormer~\cite{jiang2023polarformer} improves BEVFormer by replacing the Cartesian coordinate system with the polar coordinate system. 
Despite the unified and structured representation, BEV-based paradigms suffer from high computational cost due to the complex view transformation operations.

\paragraph{Sparse Query-based Paradigm.} 
Sparse sampling and global attention are two key branches of methods belonging to the sparse query-based paradigm.
Sparse sampling based methods suffer from limited information due to sparse sampling points and constrained receptive fields.
Following DETR3D~\cite{wang2022detr3d}, several works increase the number of sampling points and propose  approaches for the integration of their sampled features to compensate for the inadequate information resulting from single-point sampling~\cite{lin2022sparse4d,liu2023sparsebev}. 
Alternatively, global attention methods have a global receptive field, but they are also susceptible to interference from irrelevant features.
Among them, MV2D~\cite{wang2023object} generates reference points through 2D detection priors, while 3DPPE~\cite{shu20233dppe} employes 3D point encoding with the help of depth estimation.
CAPE~\cite{xiong2023cape} introduces camera view position embedding method to reduce the difficulty of view transformation learning, but it interacts with all surrounding image features through a bilateral cross-attention approach. 
To improve temporal modeling, PETRv2~\cite{liu2023petrv2} extends 3D position encoding to align 2D image features of different frames.
StreamPETR~\cite{wang2023exploring} develops an efficient object-centric temporal fusion mechanism that propagates historical 3D decoder embedding as contextual cues. 

\section{Method}

\subsection{Overall Architecture}
As illustrated in Figure~\ref{fig:framework_architecture},  $N$ surrounding images are initially processed by the 2D backbone network (e.g. ResNet~\cite{he2016deep}, VoVNet~\cite{lee2019energy}) 
to extract multi-view image features $\mathbf{F} = \{F_i \in \mathbb{R}^{C \times H_F \times W_F},i = 1, 2, \ldots , N\}$, where $C$, $H_F$ and $W_F$ represent the channel, height, and width of the feature maps, respectively.
Meanwhile, the feature maps $\mathbf{F}$ are fed into a 2D detector to obtain instance-level semantic features, which are then stored in the memory queue to enhance the object-centric temporal modeling. 

Each decoder layer consists of two key components: temporal attention and visibility cross-attention. 
In the temporal attention module, object queries interact not only with themselves but also with cached historical object-centric embedding generated from both 2D and 3D features. 
Subsequently, visibility cross-attention is independently performed in each divided local virtual space, aided by the DVPE. 
Predictions made in virtual spaces are transformed back to the global 3D world coordinate system as final detection.
The decoder embedding also updates the cached memory queue for processing future frames.

\subsection{Divided View Method}

Global cross-attention enables queries to interact with image features from all views, effectively extending the receptive field. 
However, it also induces high computational costs, some of which are unnecessary.
Specifically, queries do not need to interact with irrelevant features whose areas are far from their corresponding targets. 
Considering this, we aim to achieve the same performance as global attention by having queries interact only with the aggregated image features surrounding them. 
To minimize interference computation with view locality, the 3D world space is divided into multiple frustum-shaped spaces based on the field of view, where visibility cross-attention can be independently performed within each local space. 
Furthermore, encoding the poses and positions of different cameras through global position embedding increases the difficulties of model training.
Therefore, we transform the partitioned spaces into a unified virtual coordinate system, generating divided view position embedding. 
This enables the model to decouple position embedding from camera poses, thereby reducing the complexity of model learning. 
Finally, predictions of bounding boxes are made in the virtual coordinate system, which will be transformed back to the world coordinate system later.

\paragraph{Divided View Position Embedding.} 
We discretize the camera rays starting from each pixel at $(u, v)$ in a feature map into 3D coordinates along the predefined $D$ depth bins $\{p_d \in \mathbb{R} ^ 3, d = 1, 2, \ldots , D\}$. 
The image 3D coordinates are transformed to 3D world space as:
\begin{equation}
	p_i^{c} = H_i^{-1} p_{d,i}
\end{equation}
where $H_i \in \mathbb{R} ^ {4\times 4}$ denotes the matrix transforming the coordinates in 3D world space to the camera frustum space corresponding to the $i$-th view , and $ p_{d,i}$ represents the homogeneous coordinates of a point in the $i$-th view at the $d$-th depth. 

\begin{figure}[t]
	\centering
	\includegraphics[width=0.95\linewidth]{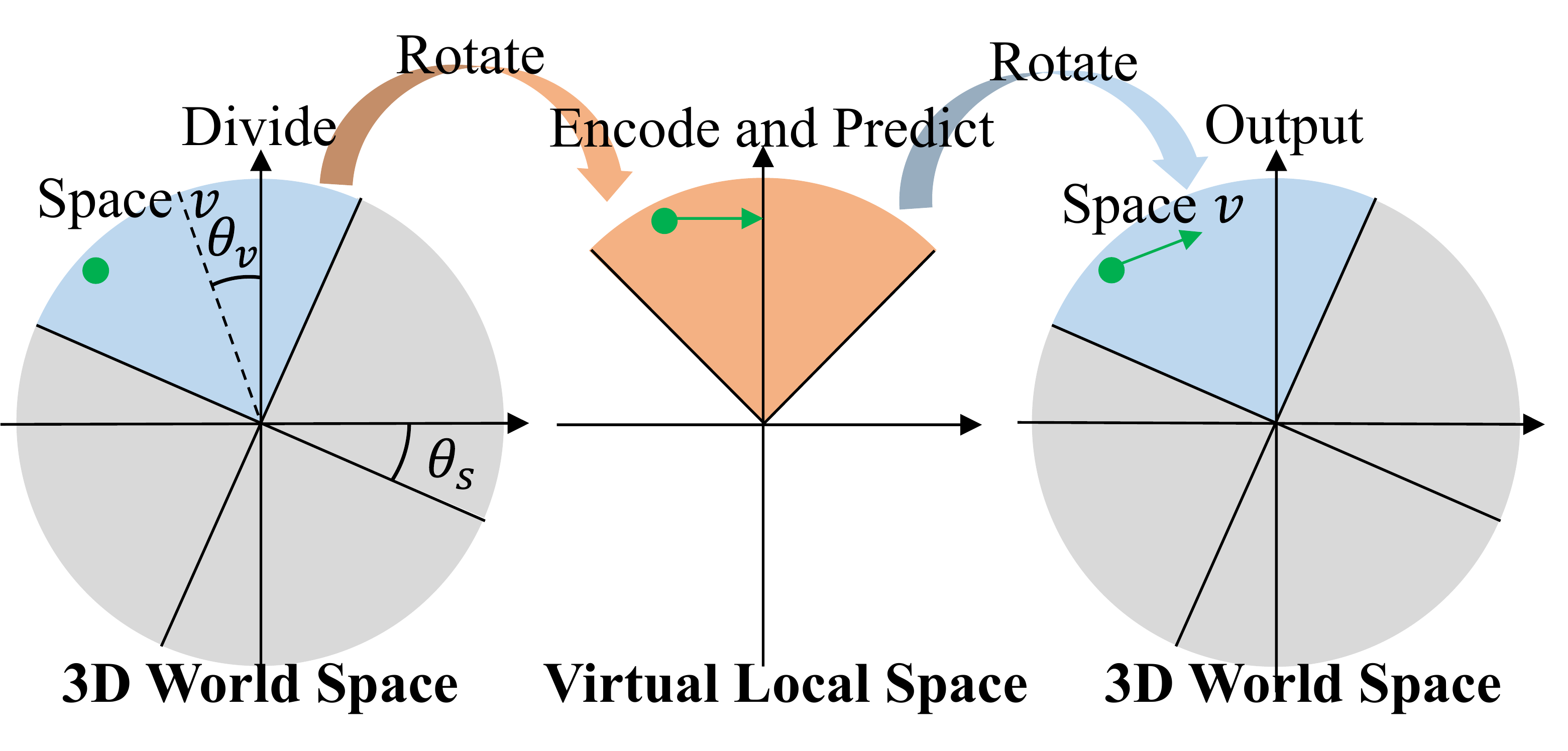}
	\caption{
        Illustration of space division (left) and transformation between the divided space and local virtual space in BEV. We only partition the global space into four, using the $v$-th space for illustration. The green dot denotes a 3D reference point, while the green arrows in the middle and on the right indicate the predicted yaws in the local virtual space and the world coordinate system.}
	\label{fig:divide_and_transform}
\end{figure}

The furthest point of each ray is selected to represent the angle (in BEV) of each image token for the subsequent view division operation.
The obtained ray-shaped camera 3D coordinates of all image tokens of $N$ views are denoted as $\mathbf P^{r} \in \mathbb{R} ^{ N \cdot H_F \cdot W_F \times D \times 3}$, while $\mathbf P^{k} \in \mathbb{R} ^{ N \cdot H_F \cdot W_F \times 3}$ represents all the furthest points of each image ray. 
Points are partitioned into $V$ groups based on their angles along the Z-axis (as illustrated on the left in Figure~\ref{fig:divide_and_transform}) as follows
\begin{equation}
    \label{divide space}
	g(p) = {\large \lfloor}  \frac{(V \times \arctan(y/x) + \theta_{s})  }{2 \pi } {\large \rfloor}  \bmod  V
\end{equation}
where $g(p)$ denotes the group index of the point $(x,y,z)$, $\theta_{s}$ is the starting shift angle (as shown in Figure \ref{fig:divide_and_transform}), and $\bmod$ represents the modulo operation. 
Based on the grouping results $g(\mathbf P^{q})$ and $g(\mathbf P^{k})$, 
3D reference points $\mathbf P^{q}$, 3D camera coordinates $\mathbf P^{r}$, 
object queries $\mathbf Q$ and flattened image features $\mathbf F$ 
are grouped into different divided views.
$M_q$ learnable 3D reference points $ \mathbf P^{q} \in \mathbb R ^{{M_q}\times 3}$ are randomly initialized within a cylindrical space to adapt to the rotational coordinate transformation needed for divided view position embedding.

Subsequently, the divided space coordinate systems are rotated along the z-axis to align with a unified virtual coordinate system with the following matrix:
\begin{equation}
    R_v = 
    \begin{bmatrix}
    \cos(\theta_v) & -\sin(\theta_v) & 0 \\
    \sin(\theta_v) & \cos(\theta_v) & 0 \\
    0 & 0 & 1
    \end{bmatrix}
\end{equation}
where $\theta_v$ is the angle difference between the $v$-th divided space and the unified virtual space. 
As illustrated from left to middle in Figure~\ref{fig:divide_and_transform},  a 3D reference point in space $v$ is transformed into the local virtual coordinate system by rotating $\theta_v$. 
Accordingly, divided view position embedding for queries and the keys, denoted as $\mathbf Q^{pe}_v$ and $\mathbf K^{pe}_v$ are obtained through respective position encoder:
\begin{align}
	\mathbf Q^{pe}_v &= \psi(R_v \mathbf P^{q}_v) \\
    \mathbf K^{pe}_v &= \xi(R_v \mathbf P^{r}_v) 
\end{align}
where $\psi$ is the query position encoding function consists of sine-cosine position
encoding function~\cite{vaswani2017attention} and multi-layer perceptron (MLP) $\xi$ 
is the position encoding function for keys as PETR~\cite{liu2022petr}.
$\mathbf P^{q}_v$ and $\mathbf P^{r}_v$ represent the 3D reference points and 3D camera coordinates of the $v$-th divided view, respectively.

\paragraph{Visibility Cross-Attention.} 
As discussed above, similar performance to global attention can be achieved when cross-attention is conducted within each local view:
\begin{equation}
    \begin{split}
	\mathbf Q_v \leftarrow \text{CrossAttn} (\text{Q} &=\mathbf{Q}_v, \  \text{K,V} = \mathbf F_v, \\
    \text{Q}^\text{PE} &=\mathbf Q^{pe}_v, \text{K}^\text{PE}=\mathbf K^{pe}_v)
    \end{split}
\end{equation}
Since the number of queries or keys in each view is not identical, we pad them to the same size for parallel computation. 
The padded invalid image features are masked during processing. 
Different views are treated as separate batches for isolation from each other. 
To ensure that the receptive field of edge queries is not affected, the starting shift angle $\theta_s$ in Eq.(\ref{divide space}) is gradually increased with different decoder layers.

\paragraph{Local Prediction Head.} 
Since queries aggregate image features in the virtual coordinate system, predictions are made in the same space as well. 
Therefore, the predicted bounding boxes need to be transformed back to the 3D world coordinate system. 
The center $c$, yaw $\alpha$, and velocity $\nu$ of the bounding box in the world space are transformed based on the rotation matrix $R_v$ and 3D reference points $\mathbf P^{q}_v$ as follows:
\begin{align}
    c =  R_v^{-1} (R_v \mathbf P^{q}_v + \hat{c}) , \quad
    \alpha =  R_v^{-1} \hat{\alpha} ,\quad \nu = R_v^{-1} \hat{\nu}
\end{align}
where $\hat{c},\hat{\alpha}, \hat{\nu}$ represent the center offset, yaw and velocity of the bounding box predicted in the local virtual space, respectively.
The transformation of yaw between two coordinate systems is shown from middle to right in Figure~\ref{fig:divide_and_transform}.

\subsection{Enhanced Object-Centric Temporal Modeling} 
Current object-centric temporal modeling typically represent historical object features with query embedding $O^{q}$ output by the decoder. 
To fully utilize the semantic information of multi-view images, we aim to incorporate historical RoI features from 2D detectors together with query embedding. 

The bounding boxes with top-k class scores are selected, of which the RoI features $\mathbf{F}^{e}$ are extracted using RoIAlign~\cite{he2017mask}.
To match the shape of query embedding, RoI features $\mathbf{F}^{e}$, together with the class predictions, are encoded as $\mathbf{O}^{e}$ using a network.
Based on the camera's intrinsic parameters $I$ and RoI features, RoI points are generated to provide positional information for RoI embedding as follows:
\begin{equation}
	\mathbf{P}^{e} = H^{-1}\text{MLP}([\text{Conv}(\mathbf{F}^{e}) ; I]).
\end{equation}

Temporal attention conducts both self-attention and temporal interaction as follows:
\begin{equation}
    \begin{split}
    \mathbf{Q} \leftarrow \text{CrossAttn}(\text{Q} &= \mathbf{Q}, \\
     \text{K, V} &= [\mathbf{Q} \  \mathbf{O}_{t-1:t-k}^{q} \  \mathbf{O}_{t-1:t-k}^{e}])
    \end{split}
\end{equation}
where $[\cdot \ \cdot]$ represents the concatenation along the sequence dimension, 
and $\mathbf{O}_{t-1:t-k}^{e}$ represents the RoI embedding of last $k$ frames. 
It is noteworthy that a standard position encoding is utilized, with cross-attention conducted in the global space. 

Following the StreamPETR~\cite{wang2023exploring}, the memory queue stores the RoI embedding $\mathbf O^{e}$ , RoI points $\mathbf P^{e}$, ego transformation matrices $\mathbf {E}$, and timestamps $\mathbf t$ for motion embedding prediction. 
The historical RoI points are transformed in response to the ego movement of the vehicle as follows (taking time $t-k$ as an example):
\begin{equation}
    \mathbf P^{e}_t = \mathbf E_{t-k}^t \mathbf P^{e}_{t-k}
\end{equation}
where $\mathbf E_{t-k}^t$ transforms the coordinates of RoI points $\mathbf P^{e}_{t-k}$ in the ego coordinate system at time $t-k$ to the one at $t$.

\subsection{One-to-Many Assignment and Loss}
DETR-based detection methods are trained using a one-to-one assignment strategy, where each ground-truth object is assigned to a single prediction. 
It results in fewer positive samples compared to the traditional paradigm, 
leading to insufficient supervision information. 
To address this issue, Group-DETR~\cite{chen2023group} introduces a one-to-many assignment approach in 2D detection to enrich the supervision for the decoder and achieves satisfactory performance. 
Therefore, we aim to explore 
the effectiveness of the one-to-many assignment strategy in 3D object detection.

Additional $G$ groups of queries are introduced in addition to the default set in the proposed framework. 
During training, queries are matched one-to-one with GTs within each group. 
These groups of queries are separated in the temporal attention module to preserve the duplicate prediction suppression ability of self-attention.  
Only the default group is activated during inference, which is identical to the models trained with the one-to-one assignment.

Unlike the default group undergoing temporal attention, the additional groups involve 
self-attention at the temporal fusion stage. 
A denoising training method~\cite{zhang2022dino} is also employed. 
The overall loss function is as follows:
\begin{equation}
    \mathcal L = \mathcal L_{2d} +  \lambda_1 \mathcal L_{3d} 
     + \lambda_2 \mathcal L_{3d}^{dn} + \lambda_3  \mathcal L_{3d}^{a}
\end{equation}
where $\mathcal L_{2d}$ denotes the 2D detection loss, $\mathcal L_{3d}$, $\mathcal L_{3d}^{dn}$ and $\mathcal L_{3d}^{a}$ represent the 3D detection loss for default queries, denoising queries and additional queries, with weight coefficients $\lambda_1$, $\lambda_2$, $\lambda_3$.
$\mathcal L_{3d}$ includes an $\ell_{1}$ loss for bounding box regression and a focal loss~\cite{lin2017focal} for classification.

\begin{table*}[t]
   \centering
   \begin{tabular}{l|ccc|cc|ccccc}  
      \toprule
      Method & Backbone & Resolution & Epochs & mAP$\uparrow$ & NDS$\uparrow$ & mATE$\downarrow$ & mASE$\downarrow$ & mAOE$\downarrow$ & mAVE$\downarrow$ & mAAE$\downarrow$ \\
      \midrule
      FB-BEV              & R50              & 704$\times$256 & 90$^\ddagger$ & 0.378 & 0.498 & 0.620 & 0.273 & 0.444 & 0.374 & 0.200 \\
      SOLOFusion      & R50              & 704$\times$256 & 90$^\ddagger$ & 0.427 & 0.534 & 0.567 & 0.274 & 0.511 & 0.252 & 0.181 \\
      Sparse4Dv2   & R50              & 704$\times$256 & 100            & 0.439 & 0.539 & 0.598 & 0.270 & 0.475 & 0.282 & 0.179 \\
      SparseBEV   & R50$^\dagger$    & 704$\times$256 & 36             & 0.448 & 0.558 & 0.581 & 0.271 & 0.373 & 0.247 & 0.190 \\
      StreamPETR & R50$^\dagger$    & 704$\times$256 & 60             & 0.450 & 0.550 & 0.613 & 0.267 & 0.413 & 0.265 & 0.196 \\
      \rowcolor{gray!20}
      \textbf{DVPE}                       & R50	            & 704$\times$256 & 90             & 0.450 & 0.555 & 0.606 & 0.271 & 0.366 & 0.269 & 0.187 \\
      \rowcolor{gray!20}
      $\textbf{DVPE}^*$                   & R50$^\dagger$   & 704$\times$256 & 60             & \textbf{0.466} & \textbf{0.559} & 0.608 & 0.271 & 0.386 & 0.274 & 0.202 \\
      \midrule
      BEVDepth      &R101             &1408$\times$512 & 90$^\ddagger$  & 0.412 & 0.535 & 0.565 & 0.266 & 0.358 & 0.331 & 0.190 \\
      BEVFormer    &R101-D$^\dagger$ &1600$\times$900 & 24             & 0.416 & 0.517 & 0.673 & 0.274 & 0.372 & 0.394 & 0.198 \\
      AeDet          &R101	            &1408$\times$512 & 90$^\ddagger$  & 0.449 & 0.561 & 0.501 & 0.262 & 0.347 & 0.330 & 0.194 \\
      SOLOFusion      &R101             &1408$\times$512 & 90$^\ddagger$  & 0.483 & 0.582 & 0.503 & 0.264 & 0.381 & 0.246 & 0.207 \\
      SparseBEV	  &R101$^\dagger$	&1408$\times$512 & 24	           & 0.501 & 0.592 & 0.562 & 0.265 & 0.321 & 0.243 & 0.195 \\
      StreamPETR &R101$^\dagger$	&1408$\times$512 & 60	           & 0.504 & 0.592 & 0.569 & 0.262 & 0.315 & 0.257 & 0.199 \\
      Sparse4Dv2   &R101$^\dagger$	&1408$\times$512 & 100	           & 0.505 & 0.594 & 0.548 & 0.268 & 0.348 & 0.239 & 0.184 \\
      \rowcolor{gray!20}
      \textbf{DVPE}                       &R101$^\dagger$	&1408$\times$512 & 60	              & \textbf{0.521} & \textbf{0.603} & 0.544 & 0.262 & 0.318 & 0.248 & 0.200 \\
      \bottomrule
   \end{tabular}
   \caption{3D object detection results on the nuScenes \texttt{val} split. $\dagger$ benefits from perspective pretraining. 
   $\ddagger$ indicates methods with CBGS which will elongate 1 epoch into 4.5 epochs. * halves the number of queries for fair comparison.}
   \label{table:nuscenes_val}
\end{table*}

\begin{table*}[t]
   \centering
   \begin{tabular}{l|cc|cc|ccccc}  
      \toprule
      Method & Backbone & Resolution & mAP$\uparrow$ & NDS$\uparrow$ & mATE$\downarrow$ & mASE$\downarrow$ & mAOE$\downarrow$ & mAVE$\downarrow$ & mAAE$\downarrow$ \\
      \midrule
      AeDet$^\dagger$ & ConvNeXt-B & 1600$\times$640 & 0.531 & 0.620 & 0.439 & 0.247 & 0.344 & 0.292 & 0.130 \\ 
      SOLOFusion & ConvNeXt-B & 1600$\times$640 & 0.540 & 0.619 & 0.453 & 0.257 & 0.376 & 0.276 & 0.148 \\ 
      StreamPETR & VoVNet-99 & 1600$\times$640 & 0.550 & 0.636 & 0.479 & 0.239 & 0.317 & 0.241 & 0.119 \\ 
      VideoBEV & ConvNeXt-B & 1600$\times$640 & 0.554 & 0.629 & 0.457 & 0.249 & 0.381 & 0.266 & 0.132 \\ 
      SparseBEV & VoVNet-99 & 1600$\times$640 & 0.556 & 0.636 & 0.485 & 0.244 & 0.332 & 0.246 & 0.117 \\ 
      Sparse4Dv2 & VoVNet-99 & 1600$\times$640 & 0.556 & 0.638 & 0.462 & 0.238 & 0.328 & 0.264 & 0.115 \\ 
      \rowcolor{gray!20}
      \textbf{DVPE}  & VoVNet-99 & 1600$\times$640  & \textbf{0.572} & \textbf{0.645} & 0.466 & 0.240 & 0.319 & 0.266 & 0.126 \\
      \bottomrule
   \end{tabular}
   \caption{3D object detection on the nuScenes \texttt{test} split. $\dagger$ uses 
   test time augmentation. All methods in the table do not use future frames.}
   \label{table:nuscenes_test}
\end{table*}

\section{Experiments}

\subsection{Dataset and Metircs}
Our framework is evaluated on the nuScenes dataset~\cite{caesar2020nuscenes}.
It contains 1k driving scenes, each with a duration of 20 seconds. 
The dataset is split into three groups: 750 for training, 150 for validation, and 150 for testing. 
Each scene includes 6 RGB images covering a 360° field 
of view, sampled at a frequency of 2Hz. 
Official metrics used in this paper include: mean Average Precision (mAP), nuScenes Detection Score (NDS), mean Average Translation Error (mATE), mean Average Scale Error (mASE), mean Average Orientation Error (mAOE), mean Average Velocity Error (mAVE) and mean Average Attribute Error (mAAE).

\subsection{Implementation Details}
We select StreamPETR~\cite{wang2023exploring} as our baseline. 
Following previous methods~\cite{liu2023sparsebev,wang2023exploring}, 
for the backbone network, we use ResNet~\cite{he2016deep} on the validation 
dataset and VoVNet~\cite{lee2019energy} on the test dataset. 
The 2D auxiliary supervision head from Focal-PETR~\cite{wang2023focal} is employed to provide 2D RoI bounding boxes. 
The AdamW~\cite{loshchilov2017decoupled} optimizer with a cosine annealing policy 
is used for training DVPE. 
The learning rate and batch size are set to $4\times10^{-4}$ and 16, respectively. 
Our models for performance comparison are trained for 60 epochs, whereas in ablation studies they are trained for 24 epochs. 
All experiments of DVPE are conducted without employing the CBGS~\cite{zhu2019class} strategy or future frames.

For the proposed framework, the 3D world space is divided into 6 spaces and the shift angle is incremented by 20 degrees at each layer. 
By default, the top 128 2D RoI features are cached in a memory queue with a length of 4 frames.  
We adopt one additional group of queries to perform one-to-many assignment training, and the number of 3D object queries and additional ones are both set to 900. 

\subsection{State-of-the-art Comparison}

\paragraph{NuScenes Val Split.} Experimental results of the proposed framework and existing methods on the validation split of nuScenes are shown in Table~\ref{table:nuscenes_val}.  When adapting ResNet50 pre-trained on ImageNet-1K~\cite{deng2009imagenet}, DVPE surpasses 
Sparse4Dv2 by 1.1\% mAP and 1.6\% NDS. When switching to backbone with nuImages pretraining, 
we reduce the number of queries to match StreamPETR and outperforms the baseline by 1.6\% 
mAP and 0.9\% NDS. With a larger backbone Resnet101 and resolution of 
1408$\times$512, DVPE also attains superior performance with 52.1\% mAP and 60.3\% NDS, 
exceeding the SOTA by 1.6\% mAP and 0.9\% NDS.
\paragraph{NuScenes Test Split.} In Table~\ref{table:nuscenes_test}, we provide the experimental results on the nuScenes test split and compare the proposed framework with previous SOTA methods . 
Following the common practice, we use VoVNet-99 as the backbone pretrained by DD3D~\cite{park2021pseudo}.
In visual 3D object detection, DVPE achieves state-of-the-art performance 
with 57.2\% mAP and 64.5\% NDS, exceeding Sparse4Dv2 by 1.6\% mAP and 0.7\% NDS.

\begin{table}[t]
    \centering
    \begin{tabular}{c|c|c|c|c|c|c}
      \toprule
      Exp & DVM & TM & OM & mAP$\uparrow$ & NDS$\uparrow$ & mATE$\downarrow$ \\
      \midrule
      a & ~ & ~ & ~ & 0.418 & 0.517 & 0.659 \\ 
      b & $\checkmark$ & ~ & ~ & 0.442 & 0.542 & 0.610 \\ 
      c & ~ & $\checkmark$ & ~ & 0.423 & 0.520 & 0.657 \\ 
      d & ~ & ~ & $\checkmark$ & 0.428 & 0.521 & 0.653 \\
      e & $\checkmark$ & $\checkmark$ & ~ & 0.446 & 0.549 & 0.604 \\ 
      f & $\checkmark$ & ~ & $\checkmark$ & 0.451 & 0.551 & \textbf{0.603} \\ 
      g & ~ & $\checkmark$ & $\checkmark$ & 0.428 & 0.518 & 0.651 \\ 
      \rowcolor{gray!20}
      h & $\checkmark$ & $\checkmark$ & $\checkmark$ & \textbf{0.455} & \textbf{0.554} & 0.604 \\
      \bottomrule
    \end{tabular}
    \caption{Ablation study on nuScenes \texttt{val} split. 
    DVM, TM and OM denote the divided view method, enhanced object-centric temporal modeling and one-to-many assignment, respectively.}
    \label{table:ablation}
\end{table}

\begin{table}[t]
    \centering
    \begin{tabular}{c|c|c|c|c}
      \toprule
      \#Views & $\theta_i$ (°) & mAP$\uparrow$ & NDS$\uparrow$ & mATE$\downarrow$      \\ 
      \midrule
      4 & 30 & 0.452 & 0.550 & 0.616   \\ 
      \rowcolor{gray!20}
      6 & 20 & \textbf{0.455} & \textbf{0.554} & 0.604\\ 
      8 & 15 & 0.453 & 0.552 & \textbf{0.601} \\ 
      \midrule
      6 & 60 & 0.449 & 0.548 & 0.614 \\ 
      6 & 30 & 0.451 & 0.549 & 0.614 \\ 
      \bottomrule
    \end{tabular}
    \caption{Results of models with different numbers of divided views and incremental steps of the shift angle $\theta_s$ at each decoder layer.}
    \label{table:divided_view}
\end{table}

\begin{figure*}[t]
	\centering
	\includegraphics[width=0.95\linewidth]{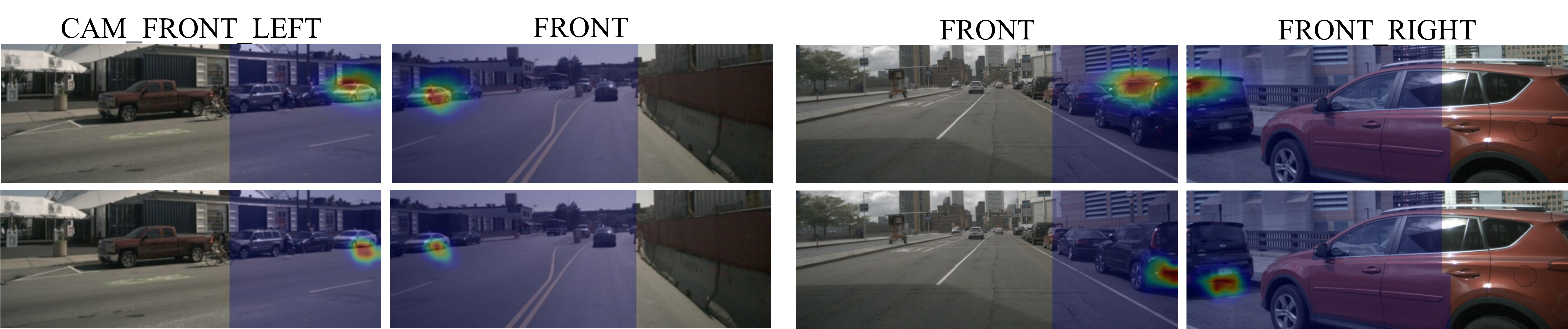}
	\caption{Visualization of image regions within one of the divided spaces and the corresponding attention maps of a query. Attention maps are from two heads of the last decoder layer.}
	\label{fig:visualization}
\end{figure*}

\begin{table}[t]
    \centering
    \begin{tabular}{l|c|c|c|c}
      \toprule
      Model & mAP$\uparrow$ & NDS$\uparrow$ & mATE$\downarrow$ & mAOE$\downarrow$ \\ 
      \midrule
      PETR & 0.299 & 0.342 & 0.766 & 0.679 \\ 
      PETR+DVM & \textbf{0.327} & \textbf{0.369} & \textbf{0.757} & \textbf{0.593} \\ 
      \bottomrule
    \end{tabular}
    \caption{Application of divided view method to PETR. Models are trained 
    for 24 epochs with Resnet50 pretrained on ImageNet-1K.}
    \label{table:divided_view_petr}
\end{table}

\begin{table}[t]
    \centering
    \begin{tabular}{c|c|c|c|c}
      \toprule
      \#RoIs & \#Frames & mAP$\uparrow$ & NDS$\uparrow$ & mATE$\downarrow$ \\ 
      \midrule
      32 & 4 & 0.450 & 0.552 & 0.616 \\ 
      64 & 4 & 0.455 & 0.553 & 0.617 \\ 
      \rowcolor{gray!20}
      128 & 4 & 0.455 & 0.554 & \textbf{0.604} \\ 
      \midrule
      128 & 2 & 0.449 & 0.550 & 0.615 \\ 
      128 & 6 & \textbf{0.456} & \textbf{0.556} & 0.614 \\ 
      \bottomrule
    \end{tabular}
    \caption{Results of experiments with different numbers of RoI proposals and cached historical frames. }
    \label{table:enhanced_object-centric_temporal_modeling}
\end{table}

\setlength{\tabcolsep}{1.5mm}
\begin{table}[t]
    \centering
    \begin{tabular}{c|c|c|c|c}

      \toprule
      \#Add'l groups & \#Add'l queries & mAP$\uparrow$ & NDS$\uparrow$ & mATE$\downarrow$ \\ 
      \midrule
      1 & 450 & 0.440 & 0.541 & 0.632 \\ 
      \rowcolor{gray!20}
      1 & 900 & 0.455 & \textbf{0.554} & \textbf{0.604} \\ 
      2 & 450 & 0.449 & 0.548 & 0.617 \\ 
      2 & 900 & \textbf{0.456} & \textbf{0.554} & 0.605 \\ 
      \bottomrule
    \end{tabular}
    \caption{Results of experiments with different numbers of additional groups and additional queries. }
    \label{table:one-to-many_assignment}
\end{table}

\subsection{Ablation Study \& Analysis}
We conduct comprehensive experiments and analyze the effectiveness of each module in our work. 
All experiments are conducted with a ResnNet50 pretrained on nuImage. 
The number of queries is set to 900, training for 24 epochs. 
Table~\ref{table:ablation} provides detailed results of ablation experiments.

\paragraph{Divided View Method.} From experiments (a) to (b) in Table~\ref{table:ablation}, 
the divided view method significantly enhances the model with increase of 2.4\% mAP and 2.5\% in 
NDS and decrease of 4.9\% mATE. In addition, the comparison between experiments (h) and (g) 
demonstrates that the removal of divided view method results in a drop of 2.7\% mAP 
and 3.6\% NDS, highlighting the method's substantial contribution to the DVPE. 
To explore the impact of the number of divided views and spatial division resulted from the shift angle, 
we conduct several experiments in Table~\ref{table:divided_view}. More variations in shift angles 
across different layers provides a more comprehensive receptive field for queries situated at the edges of the view, 
contributing to improve performance. 
\paragraph{Applicability of Divided View Method.}
We also apply the proposed DVPE to PETR, which is a single-frame framework.
As illustrated in Table~\ref{table:divided_view_petr}, the mAP and NDS are enhanced by 1.5\% and 1.2\%, respectively, and the mAOE decreases significantly by 8.6\% after incorporating our divided view method.
This indicates that our DVPE work is not limited to specific models and can be applied to many sparse query-based multi-view object detection frameworks.

\paragraph{Visualization of Divided View Method.} The visualization of image regions in a divided space and attention maps of a query is illustrated in Figure~\ref{fig:visualization}. 
Each divided space covers only a portion of six camera images. 
Instead of interacting with features from all views, the query looks up features within a divided virtual space (blue shaded regions), thus effectively reducing interference and redundant computation. 
It can be observed from the attention maps that the model can focus on targets even if they appear twice in overlapping fields of adjacent views, demonstrating the effectiveness of DVPE.
Features of targets are highlighted even when they only partially appear in both views, as it can be treated as a single view for perception with the aid of divided view position embedding, validating the remarkable localization capability of the proposed framework.

\paragraph{Enhanced Object-Centric Temporal Modeling.} Comparing experiment (a) with experiment (c) 
in Table~\ref{table:ablation}, the addition of RoI features contributes to a gain of 0.5\% mAP and 0.3\% NDS. 
This indicates that RoI features and query features are complementary, collectively providing historical cues. In Table~\ref{table:enhanced_object-centric_temporal_modeling}, we investigated the impact of the number of RoI proposals and cached historical frames on performance. 
As the number of RoI proposals increases to 128 and frames to 6, performance tends towards saturation.
\paragraph{One-to-Many Assignment.} From experiments (a) and (d) in Table~\ref{table:ablation}, 
one-to-many assignment brings an improvement of 1.0\% mAP and 0.4\% NDS. 
It demonstrates that even with denoising training, one-to-many assignment remains effective due to supplementary supervision.
Additionally, based on experiments (c), (d), and (g), joint enhancement are not obtained from the combination of temporal modeling and one-to-many assignment. 
We argue that this phenomenon can be attributed to the divergent functionalities between the temporal attention of the default queries and self-attention of the additional queries. 
The shared parameters of temporal attention may impede the capacity of model for temporal fusion.
In Table~\ref{table:one-to-many_assignment}, we further analyze the impacts of the number of additional groups and queries on the one-to-many training strategy, finding that the number of queries has a larger impact than the number of groups.

\section{Conclusion}
This paper proposes a divided view position embedding approach to effectively aggregate image features for 3D object detection.
The global 3D world space is partitioned into multiple local virtual spaces, utilizing the proposed DVPE to perceive the relative spatial locations of queries and image features. 
After subsequent visibility cross-attention, predictions are made in the local virtual coordinate system to decouple the position embedding from camera poses. 
To better utilize historical information to assist detection, we establish object-centric temporal modeling by incorporating 2D RoI features in addition to 3D query embedding.
Furthermore, we investigate the role of a one-to-many assignment training strategy in stabilizing the training of the proposed framework.
Experiments show that the DVPE achieves state-of-the-art performance on the vision-only nuScenes benchmark. 
We hope our work offers new insights into the interaction between queries and multi-view features for embodied agents.
\paragraph{Limitation and Future Work.} 
While visibility cross-attention reduces the computational cost of global attention, the frequent partition and restoration of space negate this advantage.
Therefore, there is a need for improvement to address this issue.  
Additionally, the misalignment between the one-to-many assignment and temporal attention might lead to sub-optimal results, which expects further exploration.
\section*{Acknowledgements}
This work was supported by the National Nature Science Foundation of China (Grant 62203289, 62225308), Shanghai Sailing Program (No. 22YF1413800).
\appendix

\bibliographystyle{named}
\bibliography{ijcai24}

\end{document}